\documentclass[journal,twoside,web]{ieeecolor}
\usepackage{generic}
\usepackage{cite}
\usepackage{amsmath,amssymb,amsfonts}
\usepackage{algorithmic}
\usepackage{graphicx}
\usepackage{algorithm,algorithmic}
\usepackage{hyperref}
\usepackage{amsmath}
\usepackage{multirow}

\begin{document}
\title{AI-Driven Non-Invasive Detection and Staging of Steatosis in Fatty Liver Disease Using a Novel Cascade Model and Information Fusion Techniques}

\author {{Niloufar Delfan, Pardis Ketabi Moghadam, Mohammad Khoshnevisan,  Mehdi Hosseini Chagahi, Behzad Hatami, Melika Asgharzadeh, Mohammadreza Zali, Behzad Moshiri, (\IEEEmembership{Senior member, IEEE}), Amin Momeni Moghaddam, Mohammad Amin Khalafi, Khosrow Dehnad }
\thanks{N. Delfan is with the School
of Electrical and Computer Engineering, College of Engineering, University of Tehran, Tehran, Iran, and 
Department of Electrical Engineering and Computer Science, York University, 4700 Keele Street, Toronto, ON M3J 1P3, Canada, (e-mail: ndelfan.yorku.ca).}
\thanks{M. Khoshnevisan is with Physics Department, College of Science, Northeastern University, Boston, MA 02115, USA.
Department of Computer Engineering, Antalya Bilim University, Antalya, Turkey, (m.khoshnevisan@northeastern.edu).}
\thanks{M. H. Chagahi , M. Ashgharzadeh are with the School
of Electrical and Computer Engineering, College of Engineering, University of Tehran, Tehran, Iran, (e-mail: mhdi.hoseini@ut.ac.ir).}
\thanks{P. K. Moghadam, B. Hatami, M.R. Zali, M. A. Khalafi are with the Research Institute for Gastroenterology and Liver Diseases, Shahid Beheshti University of Medical Sciences, Tehran, Iran, (e-mail: ketabimoghadam.p@gmail.com, bzd\_hatami@yahoo.com, nnzali@yahoo.com,  aminkhalafi1996@gmail.com).}
\thanks{B. Moshiri is with the School of Electrical and Computer Engineering,
College of Engineering, University of Tehran, Tehran, Iran and the Department
of Electrical and Computer Engineering University of Waterloo, Waterloo,
Canada, (e-mail: moshiri@ut.ac.ir).}
\thanks{M. A. Moghadam is with the Imaging Department, Taleghani Hospital, Shahid Beheshti University of Medical Sciences, Tehran, Iran (e-mail: draminmomeni@gmail.com).}
\thanks{K. Dehnad is with the IEOR Department, Columbia University, New York, USA (e-mail: Kd11@columbia.edu).}}
\maketitle

\begin{abstract}
Non-alcoholic fatty liver disease (NAFLD) is one of the most widespread liver disorders on a global scale, posing a significant threat of progressing to more severe conditions like nonalcoholic steatohepatitis (NASH), liver fibrosis, cirrhosis, and hepatocellular carcinoma. Diagnosing and staging NAFLD presents challenges due to its non-specific symptoms and the invasive nature of liver biopsies. Our research introduces a novel artificial intelligence cascade model employing ensemble learning and feature fusion techniques. We developed a non-invasive, robust, and reliable diagnostic artificial intelligence tool that utilizes anthropometric and laboratory parameters, facilitating early detection and intervention in NAFLD progression. Our novel artificial intelligence achieved an 86\% accuracy rate for the NASH steatosis staging task (non-NASH, steatosis grade 1, steatosis grade 2, and steatosis grade 3) and an impressive 96\% AUC-ROC for distinguishing between NASH (steatosis grade 1, grade 2, and grade3) and non-NASH cases, outperforming current state-of-the-art models. This notable improvement in diagnostic performance underscores the potential application of artificial intelligence in the early diagnosis and treatment of NAFLD, leading to better patient outcomes and a reduced healthcare burden associated with advanced liver disease.
\end{abstract}


\section{Introduction}
Nonalcoholic fatty liver disease (NAFLD) is a widespread chronic liver disorder, impacting about 25\% of the population in North America and roughly 30\% in Asia \cite{younossi2016global,chalasani2018diagnosis, kaikkonen2017metabolic}. It involves excess fat accumulation in the liver, making up at least 5\% of its weight, in individuals with minimal alcohol consumption and no other liver diseases. NAFLD can progress from nonalcoholic fatty liver (NAFL) to more severe conditions like nonalcoholic steatohepatitis (NASH), cirrhosis, and hepatocellular carcinoma (HCC) \cite{chalasani2018diagnosis}. While NAFL involves steatosis without cell damage, NASH includes inflammation and liver damage, potentially leading to cirrhosis and other complications. NASH is also linked to higher risks of cardiovascular disease and cancer due to its association with metabolic syndrome \cite{chalasani2018diagnosis}. 

Despite its health impact, NASH is underdiagnosed due to vague symptoms and limited reliable biomarkers. While liver biopsy remains the definitive diagnostic tool, noninvasive methods such as ultrasound, CT, MRI, and blood tests have been developed. Additionally, patient demographic analysis, which examines factors such as age, gender, and medical history, is used to enhance diagnostic accuracy and tailor treatment plans \cite{wong2018noninvasive}. However, these noninvasive imaging techniques come with their own set of limitations. The accuracy of ultrasound, CT scans, and MRI heavily depends on the skill and experience of the analysts interpreting the images, which can lead to variability in diagnostic results \cite{uppot2018technical}. A more practical and broadly accepted non-invasive method leverages clinical and laboratory results, particularly blood tests. Blood tests, like the NAFLD fibrosis score (NFS) and fibrosis-4 index (FIB-4), offer a practical approach to detecting advanced fibrosis with high diagnostic accuracy \cite{wong2018noninvasive}. 

Elevated levels of alanine aminotransferase (ALT) and aspartate aminotransferase (AST) are commonly observed in patients with NAFL/NASH. However, they do not correlate well with disease progression[4]. Various laboratory parameters must be combined to aid in identifying NASH, fibrosis, and steatosis. Notably, the NAFLD fibrosis score (NFS) and the fibrosis-4 index (FIB-4), recommended by the American Association for the Study of Liver Diseases (AASLD) in 2017, were commonly utilized to identify advanced fibrosis \cite{chalasani2018diagnosis}. Additional risk factors associated with NASH to liver fibrosis include fasting blood sugar (FBS), insulin resistance, hemoglobin (HB), and weight gain \cite{rinella2015nonalcoholic}. The NFS and FIB-4 are distinguished by their high diagnostic accuracy for advanced fibrosis, with AUROC values of 80-85\%, utilizing readily available clinical and biochemical parameters \cite{rinella2015nonalcoholic}.

Early intervention through lifestyle changes can reverse disease progression in patients with mild fibrosis. However, current scoring methods do not account for long-term data, limiting their accuracy \cite{rinella2015nonalcoholic, wong2018noninvasive}.  Machine learning (ML) models, incorporating clinical and laboratory data, offer a promising alternative for diagnosing and predicting disease stages \cite{adlung2021machine}. This capability allows physicians to make more informed decisions complemented by these models' insights \cite{naderi2024machine}. Various ML algorithms have been employed in the detection and classification of NAFLD, including Classification Trees \cite{zamanian2024estimation}, Random Forest (RF) \cite{garcia2021assessment}, Naïve Bayes (NB) \cite{razmpour2023application}, Neural Networks (NN) \cite{okanoue2021novel}, and Logistic Regression (LR) \cite{newsome2020fibroscan}, K‑Nearest Neighbor (KNN) \cite{razmpour2023application}, Support Vector Machine (SVM) \cite{razmpour2023application}, Adaptive Boosting (AdaBoost) \cite{razmpour2023application}, and XGBoost \cite{ji2022machine}. Clinical data used in ML models included patient demographics, electronic health records (EHRs), and blood biomarkers, which help accurately diagnose different stages of liver disease \cite{zamanian2023application}. 

However, existing studies focus on binary classification to determine the presence of NAFLD or NASH. However, accurately distinguishing between the various stages of NAFLD is equally critical. While prediction models show high accuracy when anthropometric data like waist circumference and BMI are available, their use in large-scale epidemiological research is limited due to the frequent absence of these specific parameters in many datasets. Compiling a comprehensive dataset with the necessary features and biomarkers across many samples remains a significant challenge. Therefore, developing an NAFLD prediction model that relies solely on routine clinical and laboratory parameters, which are more readily available in health databases, is essential. 

This paper introduces an innovative cascade model that utilizes ensemble learning, data, and feature fusion techniques to effectively handle missing data and enhance predictive accuracy. This ensemble-based methodology integrates diverse data sources and model predictions, ensuring that handling missing data does not compromise overall performance. As a result, our model attains superior accuracy and enhanced resilience, making it an invaluable tool for numerous applications where data quality and completeness are crucial.In summary, the primary advantages of the proposed method are:

\begin{enumerate}
  \item \textbf{Innovative Information Fusion:} The method introduces a novel stacking model that integrates data and feature fusion techniques to enhance performance.
  \item \textbf{Enhanced Data Enrichment:} The model optimizes input for the meta-classifier by combining data and feature-level information, improving its predictive capabilities.
  \item \textbf{Diverse Dataset:} The dataset, sourced from two medical centers representing rural and urban populations, ensures the model’s applicability across different demographic settings, increasing its generalizability.
  \item \textbf{Comprehensive Evaluation:} The model’s effectiveness is rigorously tested using a wide range of performance metrics and 10-fold cross-validation (CV), ensuring a thorough assessment.
  \item \textbf{Robust and Reliable System:} By leveraging multiple ML models, the approach overcomes individual model limitations, achieving 86\% accuracy in multiclass tasks and a 96\% AUC for binary classification, ensuring high reliability and performance.
  \item \textbf{Effective Handling of Missing Data:} The ensemble model efficiently manages incomplete data, a common issue in clinical settings, maintaining strong performance despite data gaps.
\end{enumerate}

\section{Methods and Materials}

\subsection{Data Acquisition and Annotation}
The Shahid Beheshti University of Medical Sciences (SBMU) ethics committee approved this study, granting it the ethical code I.R.SBMU.RIGLD.REC.1403.025. All participants provided informed consent before participating. This cross-sectional study was conducted from June 2023 to June 2024, involving 1,812 patients from the primary healthcare center in Bomehen city (a suburb of Tehran) and the family medicine group at Taleghani Hospital. These two centers were chosen to capture a diverse sample population, representing rural and urban settings, to examine differences in healthcare access, lifestyle, and metabolic diseases.

Participants included adults over 18 without underlying liver conditions, hepatobiliary cancers, excessive alcohol consumption, or comorbidities like diabetes and cardiovascular diseases. Demographic data, medical history, and lifestyle factors such as smoking and alcohol use were collected through questionnaires. Laboratory tests included measurements of various biomarkers such as ALT, AST, cholesterol levels, and the FIB-4 index. Anthropometric data, including waist-to-hip ratio and BMI, were also recorded. After a 12-hour fasting period, participants underwent laboratory tests and B-mode ultrasound to assess liver steatosis, graded from absent to severe based on liver brightness and the visibility of intrahepatic structures. Table \ref{table1} provides a detailed review of the variables available within both datasets used in the study. 

All participants underwent laboratory tests after a 12-hour fast, and patients with liver diseases were excluded based on hepatic viral marker screenings. Liver steatosis was evaluated using B-mode ultrasound to assess fatty infiltration in the liver. The ultrasound grading system was based on liver brightness, the contrast between the liver and kidney, and the visibility of intrahepatic vessels, parenchyma, and diaphragm. Liver steatosis was classified as follows: Grade 0 (absent) with normal liver echotexture; Grade 1 (mild) with slightly increased echogenicity and standard diaphragm and portal vein visibility; Grade 2 (moderate) with moderate echogenicity and slightly impaired visualization; and Grade 3 (severe) with significant echogenicity and poor or no visibility of critical structures \cite{kalfaoglu2023evaluation}.

\begin{table}[h]
\centering
\caption{Laboratory and anthropometric features of the datasets}
\begin{tabular}{|c|c|c|}
\hline
\textbf{Variable} & \textbf{Practical Definition}                                     & \textbf{Unit} \\ \hline
Age               & Age of Individuals                                               & $years$\\ \hline
Sex               & Biological Gender                                                & -             \\ \hline
WBC               & White Blood Cell Count                                           & $mm^3$             \\ \hline
HB              & Hemoglobin                                                       & $g/dL$             \\ \hline
PLT               & Platelet Count                                                   & $\mu L$\\ \hline
FIB4              & Liver Fibrosis Score                                             & -             \\ \hline
FBS               & Fasting Blood Sugar                                              & $mg/dL$            \\ \hline
AST               & Aspartate Aminotransferase                                       & $IU/L$            \\ \hline
ALT               & Alanine Aminotransferase                                         &$IU/L$            \\ \hline
Bil T             & Total Bilirubin                                                  & $mg/dL$              \\ \hline
Bil D             & Direct Bilirubin                                                 & $mg/dL$              \\ \hline
TG              & Triglyceride                                                     & $mg/dL$               \\ \hline
Chol              & Total Cholesterol                                                & $mg/dL$              \\ \hline
LDL               & Low-Density Lipoprotein                                          & $mg/dL$              \\ \hline
HDL               & High-Density Lipoprotein                                         & $mg/dL$              \\ \hline
ALB               & Albumin                                                          & $g/dL$              \\ \hline
Height            & Height of the Individual                                         & $cm$            \\ \hline
Weight            & Weight of the Individual                                         & $kg$           \\ \hline
BMI               & Body Mass Index                                                  & $kg/m^3$             \\ \hline
Waist             & Waist Circumference                                              & $cm$            \\ \hline
Hip               & Hip Circumference                                                & $cm$            \\ \hline
W/H Ratio         & Weight to Height Ratio                                           & -             \\ \hline
Fatty liver Grade & Description of the liver echogenicity  & -             \\ \hline
\end{tabular}
\label{table1}
\end{table}

\subsection{Data Preprocessing}
Since the datasets were collected retrospectively from different sources, some features, such as BMI, waist, and hip circumferences, are missing for certain subjects, leading to incomplete data. Managing missing values is crucial in data preprocessing for ML and statistical analysis, as it can affect model accuracy and lead to biased estimates. Common methods for addressing missing values include deletion, imputation, and using algorithms that can handle missing data directly \cite{nawaz2024machine}. 

Deletion removes instances or features with missing data but risks losing valuable information. Imputation fills in missing values using statistical methods like the mean, median, or more advanced techniques such as k-nearest neighbors or regression. Some ML algorithms can directly handle missing data, ensuring incomplete datasets don’t significantly impact performance \cite{nawaz2024machine}. 
To minimize biases and maintain data integrity, we created three different datasets, grouping subjects based on the available features. Categorical features were converted to numerical values in all datasets. Dataset 1 includes essential features like Age, Sex, FBS, AST, ALT, Bil T, Bil D, TG, Chol, LDL, HDL, and ALB. Dataset 2 adds WBC, HB, PLT, and FIB-4, while Dataset 3 includes Height, Weight, BMI, Waist, Hip, and W/H Ratio (Table \ref{table2}). After splitting the data, categorical features were converted to numerical values, and z-score normalization was applied to ensure consistency across datasets.

\subsection{The Proposed Model}
This project utilizes two tiers of fusion techniques: data-level fusion and feature-level fusion \cite{foo2013high,meng2020survey}. Data-level fusion involves merging raw data from various sources before any feature extraction or processing, thus capitalizing on complementary information from different sources and enhancing model performance. For instance, sensor data fusion combines information from various sensors to create a unified and complete dataset. In this project, merging laboratory test results with anthropometric metrics provides a more detailed and holistic view of a patient's condition \cite{raz2018identifying}. Feature-level fusion, or early fusion, involves extracting features independently from each data source and merging them into a single feature vector for model training. In our research, we concatenated classifier outputs and created a unified feature set for further model enhancement \cite{raz2018identifying}.

\begin{figure*}[h]
\centerline{\includegraphics[width=1.7\columnwidth]{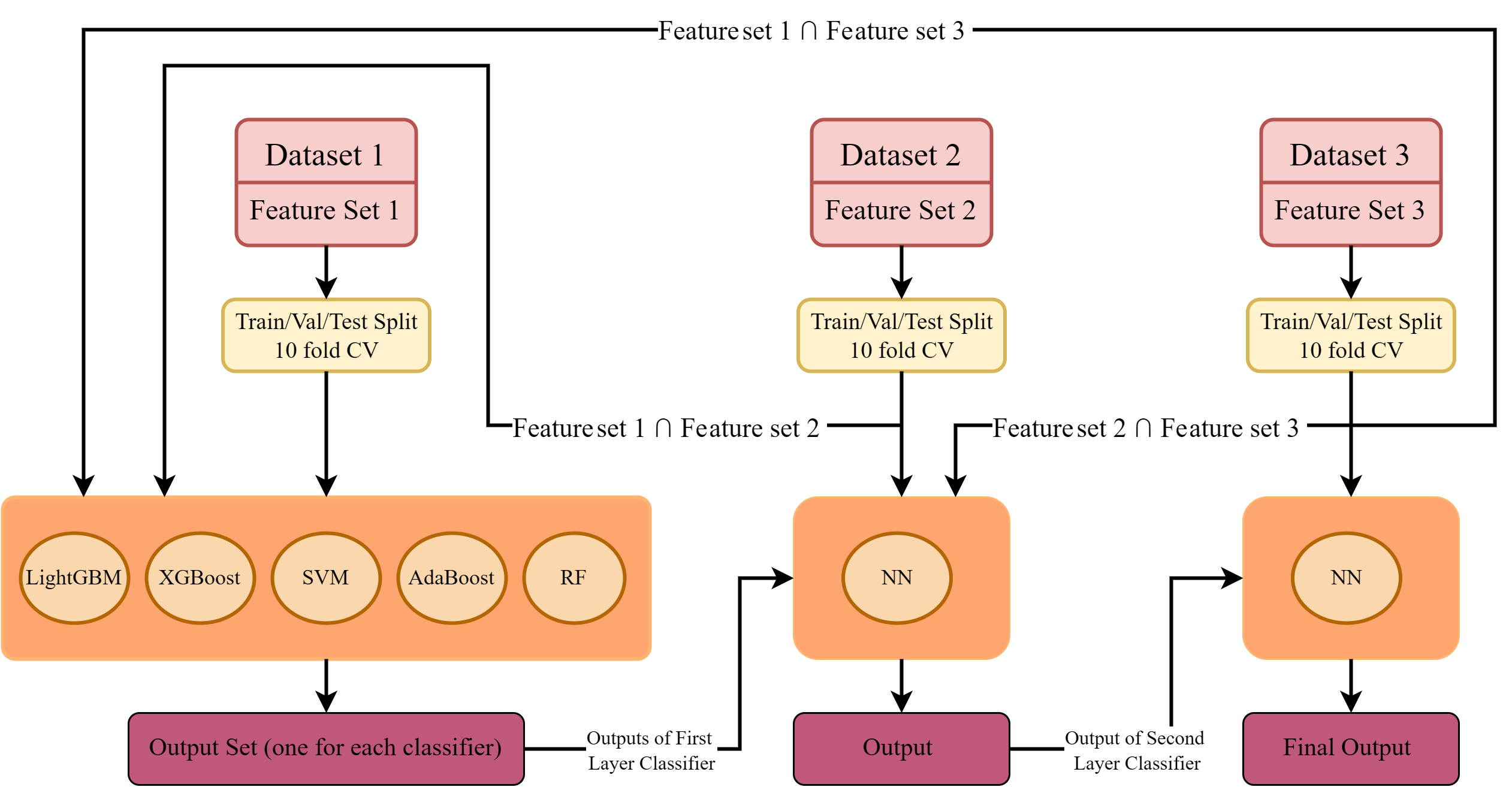}}
\caption{The proposed model architecture.}
\label{Fig1}
\end{figure*}

Fig. \ref{Fig1} depicts the architecture of the proposed model. The model architecture evaluated various classifiers, selecting the best-performing ones. Seven ML models were tested: KNN, SVM, RF, NN, AdaBoost, LightGBM, and XGBoost. SVM excels in high-dimensional spaces and resists overfitting, while RF, an ensemble of decision trees, is known for its accuracy and robustness. NN models mimic the brain's pattern recognition and handle complex data, though they require large datasets and computational resources. AdaBoost improves weak classifiers by giving more weight to misclassified instances, but it is sensitive to noise. LightGBM, optimized for large datasets, offers faster training and higher efficiency. At the same time, XGBoost, known for its flexibility and regularization, reduces overfitting and handles sparse data effectively. 

\begin{table}[t]
\centering
\caption{Clinical and laboratory features of each dataset and their characteristics}
\begin{tabular}{|c|c|c|ll}
\cline{1-3}
\textbf{Databases} & \textbf{Number of Subjects} & \textbf{Features} &  &  \\ \cline{1-3}
Database 1 &
  \begin{tabular}[c]{@{}c@{}}Grade 3: 121\\ Grade 2: 238\\ Grade 1: 149\\ Grade 0: 457\end{tabular} &
  \begin{tabular}[c]{@{}c@{}}Age, Sex, FBS, AST, ALT, \\ Bil T, Bil D, TG, Chol, LDL, \\  HDL, ALB\end{tabular} &
   &
   \\ \cline{1-3}
Database 2 &
  \begin{tabular}[c]{@{}c@{}}Grade 3: 121\\ Grade 2: 112\\ Grade 1: 127\\ Grade 0: 145\end{tabular} &
  \begin{tabular}[c]{@{}c@{}}Database 1 + \\ WBC, Hb, PLT, FIB4\end{tabular} &
   &
   \\ \cline{1-3}
Database 3 &
  \begin{tabular}[c]{@{}c@{}}Grade 3: 164\\ Grade 2: 70\\ Grade 1: 58\\ Grade 0: 50\end{tabular} &
  \begin{tabular}[c]{@{}c@{}}Database 2 + \\ Height, Weight, \\ BMI, Waist, Hip, WHR\end{tabular} &
   &
   \\ \cline{1-3}
\end{tabular}
\label{table2}
\end{table}

The proposed model utilizes a sophisticated three-layer stacked ensemble learning framework, integrating multiple datasets and feature sets to enhance prediction accuracy. In the first layer, a feature stacking technique is employed. Feature stacking involves training various models on the original feature sets and then using the predictions from these models as input features for the next layer model. This approach helps capture more complex relationships between the features. Accordingly, a diverse set of classifiers— KNN, SVM, RF, NN, AdaBoost, LightGBM, and XGBoost are trained on Dataset 1 using Feature-set 1 $(F^1)$ (Equation \ref{eq1}). The training employed a 10-fold CV to ensure robust performance and mitigate overfitting. Classifiers with an average test accuracy exceeding 70\% were selected, resulting in a final set of five classifiers (SVM, RF, AdaBoost, LightGBM, and XGBoost). Each classifier generates an output, forming an output set that captures diverse perspectives from different algorithms (Equation \ref{eq2}).\\

\textit{For first layer training:}
\begin{equation}
\text{Input (Layer 1)} : I_1 =  \{F^1_1, F^1_2, \dots, F^1_n\}
\label{eq1}
\end{equation}

\begin{equation}
    \begin{split}
        &\text{Output (Layer 1)} : O_1 =  \{SVM(I_1), RF(I_1), \\
        & XGBoost(I_1), LightGBM(I_1), AdaBoost(I_1)\}
    \end{split}
\label{eq2}
\end{equation}

For training the second-layer classifier, first, the intersection of Feature-set 1 $(F^1)$ and Feature-set 2 $(F^2)$ was fed into the first-layer classifiers (Equation \ref{eq3}).  The outputs from these classifiers and Feature-set 2 were combined and fed into the second-layer classifier (Equations \ref{eq4}, \ref{eq5}, \ref{eq6}). This classifier synthesized the information from the diverse models and Feature-set 2 $(F^2)$ to make second-layer predictions. The output from this layer represents an integrated prediction, combining the strengths of various models and feature sets. All seven base classifiers were also tested for the second layer, with the NN showing the highest average accuracy and thus being selected. \\

\textit{For second layer training:}
\begin{equation}
\text{Input (Layer 1)} : I'_1 = \{F_i \mid F_i \in ({F^1} \cap {F^2})\}
\label{eq3}
\end{equation}

\begin{equation}
    \begin{split}
        &\text{Output (Layer 1)} : O'_1 =  \{SVM(I'_1), RF(I'_1), \\
        & XGBoost(I'_1), LightGBM(I'_1), AdaBoost(I'_1)\}
    \end{split}
\label{eq4}
\end{equation}

\begin{equation}
\text{Input (Layer 2)} : I'_2 = \{F_i \mid F_i \in ({F^2} || {O'_1})\}
\label{eq5}
\end{equation}

\begin{equation}
    \begin{split}
        &\text{Output (Layer 2)} : O'_2 =  \{NN(I'_2)\}
    \end{split}
\label{eq6}
\end{equation}

In the third layer (meta-classifier), the intersection of Feature-set 1 $(F^1)$ and Feature-set 3 $(F^3)$ was processed by the first-layer classifiers (Equations \ref{eq7} and \ref{eq8}). The outputs from these classifiers and the intersection of Feature-set 2 $(F^2)$ and Feature-set 3 $(F^3)$ was combined and input into the second-layer classifier (Equations \ref{eq9} and \ref{eq10}). The output of the second layer was then merged with Feature-set 3 $(F^3)$ and fed into the third-layer classifier (Equation \ref{eq11}). This final classifier processed the combined information to produce the final prediction (Equation \ref{eq12}). Again, all seven base classifiers were tested for this layer, with NN achieving the highest average accuracy and being selected.\\

\textit{For third layer training:}
\begin{equation}
\text{Input (Layer 1)} : I^"_1 = \{F_i \mid F_i \in ({F^1} \cap {F^3})\}
\label{eq7}
\end{equation}

\begin{equation}
    \begin{split}
        &\text{Output (Layer 1)} : O^"_1 =  \{SVM(I^"_1), RF(I^"_1), \\
        & XGBoost(I^"_1), LightGBM(I^"_1), AdaBoost(I^"_1)\}
    \end{split}
\label{eq8}
\end{equation}

\begin{equation}
\text{Input (Layer 2)} : I^"_2 = \{F_i \mid F_i \in ({F^2} \cap {F^3} || {O^"_1})\}
\label{eq9}
\end{equation}

\begin{equation}
    \begin{split}
        &\text{Output (Layer 2)} : O^"_2 =  \{NN(I^"_2)\}
    \end{split}
\label{eq10}
\end{equation}

\begin{equation}
\text{Input (Layer 3)} : I^"_3 = \{F_i \mid F_i \in ({F^3} || {O^"_2})\}
\label{eq11}
\end{equation}

\begin{equation}
    \begin{split}
        &\text{Output (Layer 3)} : O^"_3 =  \{NN(I^"_3)\}
    \end{split}
\label{eq12}
\end{equation}

This ensemble approach is especially beneficial for complex prediction tasks where each model captures unique data aspects. By stacking and integrating these models, the proposed framework attains greater accuracy and reliability, taking advantage of the collective consensus of the top-performing classifiers.

\begin{table*}[]
\centering
\caption{Models hyperparameters, their search intervals, and their optimum value.}
\begin{tabular}{|c|c|c|}
\hline
\textbf{Databases} &
\textbf{Hyperparameters and their Search intervals}  &
\textbf{Optimal Values} \\ \hline
SVM &
  \begin{tabular}[c]{@{}c@{}}'C': {[}0.001, 0.01, 0.1, 1, 10, 100{]}\\ 'kernel': {[}'linear', 'poly', 'rbf', 'sigmoid'{]}\end{tabular} &
  \begin{tabular}[c]{@{}c@{}}'C': 0.1\\ 'kernel': 'rbf'\end{tabular} \\ \hline
RF &
  \begin{tabular}[c]{@{}c@{}}'n\_estimators': {[}100, 200, 300, 400, 500{]}\\ 'max\_depth': {[}None, 10, 20, 30, 40, 50{]}\\ 'min\_samples\_split': {[}2, 5, 10, 15, 20{]}\\ 'min\_samples\_leaf': {[}1, 2, 4, 6, 8{]}\\ 'max\_features': {[}'auto', 'sqrt', 'log2'{]}\end{tabular} &
  \begin{tabular}[c]{@{}c@{}}'n\_estimators': 400\\ 'max\_depth': 30\\ 'min\_samples\_split':  5\\ 'min\_samples\_leaf':  4\\ 'max\_features': 'sqrt'\end{tabular} \\ \hline
LightGBM &
  \begin{tabular}[c]{@{}c@{}}'n\_estimators': {[}10, 20, 40, 50, 100, 500, 1000, 2000, 5000{]}\\ 'num\_leaves': sp\_randint (6, 50)\\ 'learning\_rate': {[}0.0001,0.001,0.01, 0.1, 1{]}\\  'min\_child\_samples': sp\_randint (100, 500)\\  'min\_child\_weight': {[}1e-5, 1e-3, 1e-2, 1e-1, 1, 1e1, 1e2, 1e3, 1e4{]}\\ 'subsample': sp\_uniform(loc=0.2, scale=0.8)\\ 'colsample\_bytree': sp\_uniform(loc=0.4, scale=0.6)\\ 'reg\_alpha': {[}0, 1e-1, 1, 2, 5, 7, 10, 50, 100{]}\\  'reg\_lambda': {[}0, 1e-1, 1, 5, 10, 20, 50, 100{]}\end{tabular} &
  \begin{tabular}[c]{@{}c@{}}'n\_estimators': 1000\\ 'num\_leaves': 40\\ 'learning\_rate':  0.1\\  'min\_child\_samples': 207\\  'min\_child\_weight':  1e-3\\ 'subsample':  0.200190018860\\ 'colsample\_bytree': 0.7121008659\\ 'reg\_alpha': 1e-1\\ 'reg\_lambda': 20\end{tabular} \\ \hline
AdaBoost &
  \begin{tabular}[c]{@{}c@{}}'n\_estimators' : {[}10, 20, 40, 50, 100, 500, 1000, 2000, 5000{]}\\ 'learning\_rate' : {[}0.001, 0.01, 0.1, 0.5, 1{]}\\ 'Base\_estimator\_max\_depth' : {[}1, 2, 3, 4, 5, 6, 7, 8, 9, 10{]}\end{tabular} &
  \begin{tabular}[c]{@{}c@{}}'n\_estimators' : 500\\ 'learning\_rate' : 0.1\\ 'Base\_estimator\_max\_depth' :  4\end{tabular} \\ \hline
XGBoost &
  \begin{tabular}[c]{@{}c@{}}'n\_estimators': {[}10, 20, 40, 50, 100, 500, 1000, 2000, 5000{]}\\ 'learning\_rate': {[}0.001, 0.01, 0.1, 0.3, 0.5, 1{]}\\ 'max\_depth': {[}3, 4, 5, 6, 7, 8, 9, 10{]}\\ 'subsample': {[}0.6, 0.7, 0.8, 0.9, 1.0{]}\\ 'colsample\_bytree': {[}0.6, 0.7, 0.8, 0.9, 1.0{]}\\ 'gamma': {[}0, 0.1, 0.2, 0.3, 0.4, 0.5{]}\end{tabular} &
  \begin{tabular}[c]{@{}c@{}}'n\_estimators': 2000\\ 'learning\_rate': 0.01\\ 'max\_depth':  6\\ 'subsample': 0.8\\ 'colsample\_bytree': 0.6\\ 'gamma': 0.3\end{tabular} \\ \hline
\begin{tabular}[c]{@{}c@{}}NN\\ (Second-layer)\end{tabular} &
  \multirow{2}{*}{\begin{tabular}[c]{@{}c@{}}'hidden\_layer\_sizes': {[}(50,), (100,), (50, 50), (100, 100), \\ (100,50), (100, 50, 100), (100,100,50){]}\\ 'activation': {[}'tanh', 'relu', 'logistic'{]}\\ 'solver': {[}'sgd', 'adam'{]}\\ 'alpha': {[}0.0001, 0.001, 0.01, 0.1{]}\\ 'learning\_rate': {[}'constant', 'invscaling', 'adaptive'{]}\\ 'learning\_rate\_init': {[}0.001, 0.01, 0.1{]}\\ 'batch\_size': {[}32, 64, 128{]}\\ 'momentum': {[}0.9, 0.95, 0.99{]}\end{tabular}} &
  \begin{tabular}[c]{@{}c@{}}'hidden\_layer\_sizes': (100, 50, 100)\\ 'activation': 'relu'\\ 'solver': 'adam'\\ 'alpha': 0.001\\ 'learning\_rate': 'adaptive'\\ 'learning\_rate\_init': 0.001\\ 'batch\_size': 32\\ 'momentum': 0.9\end{tabular} \\ \cline{1-1} \cline{3-3} 
\begin{tabular}[c]{@{}c@{}}NN\\ (Third-layer)\end{tabular} &
   &
  \begin{tabular}[c]{@{}c@{}}'hidden\_layer\_sizes': (100,100,50)\\ 'activation': 'relu'\\ 'solver': 'sgd'\\ 'alpha': 0.001\\ 'learning\_rate': 'adaptive'\\ 'learning\_rate\_init': 0.001\\ 'batch\_size': 32\\ 'momentum': 0.9\end{tabular} \\ \hline
\end{tabular}
\label{table3}
\end{table*}

\subsection{Cross Validation, Hyperparameter Tuning, and Model Training}
10-fold CV is commonly used to evaluate a model’s performance and generalization ability. It divides the dataset into ten equal parts, using one part as the test set and the remaining nine for training. This process repeats ten times, with each fold serving as the test set once, ensuring comprehensive data use for training and testing. Hyperparameter tuning optimizes model parameters for better learning and is integrated with a 10-fold CV \cite{bischl2023hyperparameter}. This study used a random search approach to explore a wide range of hyperparameter values, improving the chances of finding the best settings \cite{bischl2023hyperparameter}. The random search strategy allowed for a broad and efficient exploration of the hyperparameter space, enhancing the likelihood of identifying each model's most effective parameter settings. 

Table \ref{table3} presents the model's hyperparameters, search ranges, and optimal values. Each model underwent independent tuning, with the best-performing classifier for each layer selected based on its metrics. Additionally, hyperparameter tuning was meticulously performed for each model to ensure a thorough and fair evaluation. This careful tuning process is crucial in optimizing the models' performance, providing a more reliable assessment of their capabilities. The combination of 10-fold CV and random search for hyperparameter tuning was designed to build a reliable, accurate model with strong generalizability, essential for real-world application.

\section {Results}           

\subsection{Model statistics}
The effectiveness of this method is evaluated using performance metrics such as accuracy (Acc), sensitivity (Sens), and specificity (Spec). Accuracy is the ratio of correctly classified cases (true positives and true negatives) to the total number of cases. Sensitivity measures the proportion of true positives out of all actual positives, while specificity is the proportion of true negatives out of all actual negatives. Additionally, the model's performance is summarized with the area under the ROC curve (AUC-ROC), where values closer to 1 indicate better performance. These metrics comprehensively assess the model's accuracy, reliability, and robustness in diagnosing and predicting outcomes.

\subsection{Experiment results}
\begin{figure}[t]
\centerline{\includegraphics[width=\columnwidth]{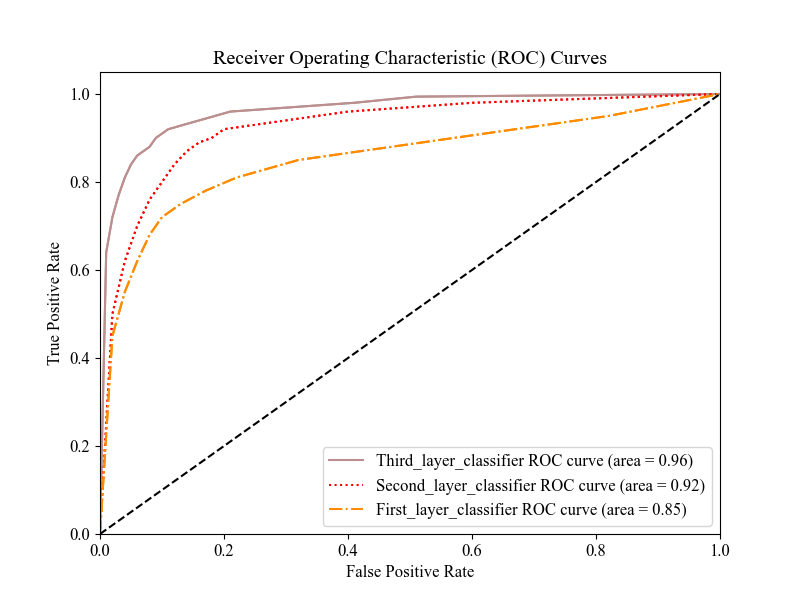}}
\caption{AUC-ROC curve for prediction of fatty liver.}
\label{Fig2}
\end{figure}

The proposed method for detecting and staging NAFLD was evaluated using performance metrics such as accuracy, sensitivity, and specificity to assess predictive accuracy and reliability. Multiple ML models—LightGBM, XGBoost, AdaBoost, SVM, RF, and NN—were tested using 10-fold CV. A three-layer stacking model was developed, with each layer trained and validated independently. Table \ref{table4} summarizes the performance of each model, showing improved results with each layer. The third layer achieved the best performance, with an accuracy of 86.9\%, sensitivity of 87.3\%, and specificity of 95.9\%. The AUC-ROC was plotted to compare the performance of various classification models. Fig. \ref{Fig2} displays the ROC curves at different levels, providing a summary of their comparative effectiveness.

\begin{table}[]
\centering
\caption{Performance measures were obtained using different model architectures with a 10-fold CV. All performance metrics are presented in percentage (\%).}
\begin{tabular}{|c|c|cccc|}
\hline
\textbf{Model} &
  \textbf{Metrics} &
  \multicolumn{1}{c|}{\textbf{\begin{tabular}[c]{@{}c@{}}Grade \\ 3\end{tabular}}} &
  \multicolumn{1}{c|}{\textbf{\begin{tabular}[c]{@{}c@{}}Grade \\ 2\end{tabular}}} &
  \multicolumn{1}{c|}{\textbf{\begin{tabular}[c]{@{}c@{}}Grade \\ 1\end{tabular}}} &
  \textbf{\begin{tabular}[c]{@{}c@{}}Grade\\ 0\end{tabular}} \\ \hline
\multirow{6}{*}{LightGBM} &
  Sens &
  \multicolumn{1}{c|}{64.4} &
  \multicolumn{1}{c|}{68.4} &
  \multicolumn{1}{c|}{71.8} &
  78.4 \\ \cline{2-6} 
 &
  Spec &
  \multicolumn{1}{c|}{92.9} &
  \multicolumn{1}{c|}{91.3} &
  \multicolumn{1}{c|}{86.6} &
  94.7 \\ \cline{2-6} 
 &
  M F1-Score &
  \multicolumn{4}{c|}{68.5} \\ \cline{2-6} 
 &
  Acc &
  \multicolumn{4}{c|}{73.2} \\ \cline{2-6} 
 &
  M-avg Sens &
  \multicolumn{4}{c|}{70.7} \\ \cline{2-6} 
 &
  M-avg Spec &
  \multicolumn{4}{c|}{91.4} \\ \hline
\multirow{6}{*}{XGBoost} &
  Sens &
  \multicolumn{1}{c|}{63.6} &
  \multicolumn{1}{c|}{69.2} &
  \multicolumn{1}{c|}{72.4} &
  78.6 \\ \cline{2-6} 
 &
  Spec &
  \multicolumn{1}{c|}{92.9} &
  \multicolumn{1}{c|}{91.6} &
  \multicolumn{1}{c|}{86.5} &
  95.1 \\ \cline{2-6} 
 &
  M F1-Score &
  \multicolumn{4}{c|}{68.7} \\ \cline{2-6} 
 &
  Acc &
  \multicolumn{4}{c|}{73.5} \\ \cline{2-6} 
 &
  M-avg Sens &
  \multicolumn{4}{c|}{71.0} \\ \cline{2-6} 
 &
  M-avg Spec &
  \multicolumn{4}{c|}{91.5} \\ \hline
\multirow{6}{*}{AdaBoost} &
  Sens &
  \multicolumn{1}{c|}{66.1} &
  \multicolumn{1}{c|}{69.2} &
  \multicolumn{1}{c|}{72.4} &
  79.0 \\ \cline{2-6} 
 &
  Spec &
  \multicolumn{1}{c|}{93.5} &
  \multicolumn{1}{c|}{91.7} &
  \multicolumn{1}{c|}{86.6} &
  94.7 \\ \cline{2-6} 
 &
  M F1-Score &
  \multicolumn{4}{c|}{69.5} \\ \cline{2-6} 
 &
  Acc &
  \multicolumn{4}{c|}{74.0} \\ \cline{2-6} 
 &
  M-avg Sens &
  \multicolumn{4}{c|}{71.7} \\ \cline{2-6} 
 &
  M-avg Spec &
  \multicolumn{4}{c|}{91.6} \\ \hline
\multirow{6}{*}{SVM} &
  Sens &
  \multicolumn{1}{c|}{61.9} &
  \multicolumn{1}{c|}{65.7} &
  \multicolumn{1}{c|}{70.4} &
  76.6 \\ \cline{2-6} 
 &
  Spec &
  \multicolumn{1}{c|}{91.8} &
  \multicolumn{1}{c|}{90.2} &
  \multicolumn{1}{c|}{86.0} &
  95.3 \\ \cline{2-6} 
 &
  M F1-Score &
  \multicolumn{4}{c|}{66.3} \\ \cline{2-6} 
 &
  Acc &
  \multicolumn{4}{c|}{71.2} \\ \cline{2-6} 
 &
  M-avg Sens &
  \multicolumn{4}{c|}{68.7} \\ \cline{2-6} 
 &
  M-avg Spec &
  \multicolumn{4}{c|}{90.8} \\ \hline
\multirow{6}{*}{RF} &
  Sens &
  \multicolumn{1}{c|}{61.9} &
  \multicolumn{1}{c|}{68.4} &
  \multicolumn{1}{c|}{70.4} &
  78.0 \\ \cline{2-6} 
 &
  Spec &
  \multicolumn{1}{c|}{92.7} &
  \multicolumn{1}{c|}{91.2} &
  \multicolumn{1}{c|}{86.8} &
  93.5 \\ \cline{2-6} 
 &
  M F1-Score &
  \multicolumn{4}{c|}{67.6} \\ \cline{2-6} 
 &
  Acc &
  \multicolumn{4}{c|}{72.5} \\ \cline{2-6} 
 &
  M-avg Sens &
  \multicolumn{4}{c|}{69.7} \\ \cline{2-6} 
 &
  M-avg Spec &
  \multicolumn{4}{c|}{91.1} \\ \hline
\multirow{6}{*}{\begin{tabular}[c]{@{}c@{}}First\_layer\\ Classifier\end{tabular}} &
  Sens &
  \multicolumn{1}{c|}{72.7} &
  \multicolumn{1}{c|}{70.7} &
  \multicolumn{1}{c|}{74.6} &
  80.3 \\ \cline{2-6} 
 &
  Spec &
  \multicolumn{1}{c|}{93.8} &
  \multicolumn{1}{c|}{93.2} &
  \multicolumn{1}{c|}{87.3} &
  94.8 \\ \cline{2-6} 
 &
  M F1-Score &
  \multicolumn{4}{c|}{72.3} \\ \cline{2-6} 
 &
  Acc &
  \multicolumn{4}{c|}{76.1} \\ \cline{2-6} 
 &
  M-avg Sens &
  \multicolumn{4}{c|}{74.6} \\ \cline{2-6} 
 &
  M-avg Spec &
  \multicolumn{4}{c|}{92.3} \\ \hline
\multirow{6}{*}{\begin{tabular}[c]{@{}c@{}}Second-layer\\ Classifier\end{tabular}} &
  Sens &
  \multicolumn{1}{c|}{77.6} &
  \multicolumn{1}{c|}{76.8} &
  \multicolumn{1}{c|}{82.6} &
  82.2 \\ \cline{2-6} 
 &
  Spec &
  \multicolumn{1}{c|}{95.3} &
  \multicolumn{1}{c|}{94.5} &
  \multicolumn{1}{c|}{88.8} &
  96.4 \\ \cline{2-6} 
 &
  M F1-Score &
  \multicolumn{4}{c|}{77.4} \\ \cline{2-6} 
 &
  Acc &
  \multicolumn{4}{c|}{80.4} \\ \cline{2-6} 
 &
  M-avg Sens &
  \multicolumn{4}{c|}{79.8} \\ \cline{2-6} 
 &
  M-avg Spec &
  \multicolumn{4}{c|}{93.7} \\ \hline
\multirow{6}{*}{\begin{tabular}[c]{@{}c@{}}Third-layer\\ Classifier\end{tabular}} &
  Sens &
  \multicolumn{1}{c|}{85.1} &
  \multicolumn{1}{c|}{86.8} &
  \multicolumn{1}{c|}{91.3} &
  85.9 \\ \cline{2-6} 
 &
  Spec &
  \multicolumn{1}{c|}{97.4} &
  \multicolumn{1}{c|}{96.2} &
  \multicolumn{1}{c|}{91.4} &
  98.4 \\ \cline{2-6} 
 &
  M F1-Score &
  \multicolumn{4}{c|}{85.0} \\ \cline{2-6} 
 &
  Acc &
  \multicolumn{4}{c|}{86.9} \\ \cline{2-6} 
 &
  M-avg Sens &
  \multicolumn{4}{c|}{87.3} \\ \cline{2-6} 
 &
  M-avg Spec &
  \multicolumn{4}{c|}{95.9} \\ \hline
\end{tabular}
\label{table4}
\end{table}

\section{Discussion}
The study presents a significant advancement in NAFLD diagnosis using ensemble learning techniques. The proposed multi-layer stacking model outperforms conventional models, particularly in sensitivity and specificity, essential for accurately detecting the disease. By leveraging multiple ML algorithms and integrating diverse data sources, the model uses advanced information fusion techniques to enhance its predictive capabilities.

The model's design incorporates advanced information fusion techniques, enhancing its ability to utilize available data comprehensively. This ensures a more robust and accurate diagnostic tool. One key advantage of this methodology is its effective handling of missing data. In real-world clinical settings, incomplete datasets are common and can undermine the performance of traditional models. The multi-layer stacking model addresses this issue by maintaining overall performance even when some features are unavailable. This improves the reliability of diagnostic predictions and increases the model's resilience and adaptability to varying data conditions.

\begin{table*}[h]
\centering
\caption{Summary of comparison of our proposed method and other methods.}
\begin{tabular}{|c|c|c|c|c|c|}
\hline
\textbf{Work} &
  \textbf{Data size} &
  \textbf{Features} &
  \textbf{TASK} &
  \textbf{ML algorithm} &
  \textbf{\begin{tabular}[c]{@{}c@{}}Performance \\ metrics\end{tabular}} \\ \hline
{[}10{]} &
  1525 &
  \begin{tabular}[c]{@{}c@{}}Sex, Age, Height, Weight, SBP, DBP, \\ Chol, FBS, glycated HB, Insulin, vitamin D\end{tabular} &
  \begin{tabular}[c]{@{}c@{}}NASH vs. \\  no NASH\end{tabular} &
  RF &
  \begin{tabular}[c]{@{}c@{}}AUC = 83\\ Acc = 87\\ Sens = 64\\ Spec = 96\end{tabular} \\ \hline
{[}26{]} &
  10,508 &
  \begin{tabular}[c]{@{}c@{}}Sex, BMI, Age, ALT, AST. Alkaline phosphatase, \\ GGT, Bil T, Bil D, Chol, TG, LDL, HDL, FBS, Uric acid\end{tabular} &
  \begin{tabular}[c]{@{}c@{}}NASH vs.  \\ no NASH\end{tabular} &
  BN &
  \begin{tabular}[c]{@{}c@{}}Acc = 83\\ Sens = 67\\ Spec = 87\end{tabular} \\ \hline
{[}25{]} &
  704 &
  \begin{tabular}[c]{@{}c@{}}Age, Height, BMI, AST, ALT, AST/ALT, ALB, TG, \\ HbA1C, Total protein, PLT, WBC, Hypertension, Hematocrit\end{tabular} &
  \begin{tabular}[c]{@{}c@{}}NASH vs.  \\ no NASH\end{tabular} &
  XGBoost &
  \begin{tabular}[c]{@{}c@{}}AUC = 82\\ Sens = 81\end{tabular} \\ \hline
{[}27{]} &
  281 &
  \begin{tabular}[c]{@{}c@{}}Age, Height, BMI, AST, ALT, AST/ALT, ALB, TG, \\ HbA1C, Total protein, PLT, WBC, Hypertension, Hematocrit\end{tabular} &
  \begin{tabular}[c]{@{}c@{}}NASH vs.  \\ no NASH\end{tabular} &
  XGBoost &
  \begin{tabular}[c]{@{}c@{}}Acc = 75\\ Sens = 81\\ Spec = 63\end{tabular} \\ \hline
{[}9{]} &
  181 &
  \begin{tabular}[c]{@{}c@{}}BMI, Sex, Age, Lipid-lowering, agents, ALT,\\ creatinine, glycated HB, Chol, TG, \\ HDL, LDL, lipoprotein(a), loge(lipoprotein(a))\end{tabular} &
  \begin{tabular}[c]{@{}c@{}}NASH vs.  \\ no NASH\end{tabular} &
  \begin{tabular}[c]{@{}c@{}}LR, RF, \\ AdaBoost,\\ KNN, SVM, \\ MLP, DT\end{tabular} &
  \begin{tabular}[c]{@{}c@{}}AUC = 79\\ Acc = 81\end{tabular} \\ \hline
{[}11{]} &
  513 &
  \begin{tabular}[c]{@{}c@{}}Abdomen circumference, Waist, \\ Chest circumference, Trunk fat, BMI\end{tabular} &
  \begin{tabular}[c]{@{}c@{}}Steatosis\\ Staging\end{tabular} &
  \begin{tabular}[c]{@{}c@{}}SVM with RBF,\\  GP, RF, NN, KNN,\\ AdaBoost, NB\end{tabular} &
  Acc = 52 \\ \hline
\begin{tabular}[c]{@{}c@{}}The proposed \\ model\end{tabular} &
  1812 &
  \begin{tabular}[c]{@{}c@{}}Age, Sex, WBC, Hb, PLT, FIB4, FBS, \\ AST, ALT, Bil T, Bil D, T.G., Chol, LDL, \\ HDL, Alb, Height, Weight, BMI, Waist, Hip, WHR\end{tabular} &
  \begin{tabular}[c]{@{}c@{}}Steatosis\\ Staging\end{tabular} &
  Ensemble Learning &
  \begin{tabular}[c]{@{}c@{}}Acc = 86\\ Sens = 87\\ Spec = 95\end{tabular} \\ \hline
\end{tabular}
\label{table5}
\end{table*}

The Use of information fusion and ensemble learning in healthcare has demonstrated substantial promise for improving diagnostic accuracy, forecasting patient outcomes, and customizing treatment plans. Data fusion methods combine various data sources, including EHRs, imaging data, genomics, and sensor data, to offer a holistic view of a patient's health status. This holistic approach enables more precise and timely interventions. Recent studies have demonstrated the efficacy of combining data fusion with ensemble learning to enhance the robustness and generalizability of predictive models in intricate clinical settings \cite{esteva2019guide}. For example, the fusion of clinical and imaging data using ensemble techniques has enhanced the early diagnosis of diseases such as cancer and cardiovascular conditions \cite{chagahi2024cardiovascular, dehghan2023breast, anstee2013progression}. Previously, Authors of \cite{docherty2021development} developed an ensemble learning framework for detecting all-cause advanced hepatic fibrosis. Nonetheless, as far as we know, this research is the first to integrate information fusion techniques and ensemble learning for staging liver steatosis.

Table \ref{table5} extensively reviews multiple studies employing ML to diagnose NASH-related diseases using clinical data. The studies are categorized based on the classification tasks they address, the features they use, their ML methods, and the number of subjects involved. To distinguish NASH cases from non-NASH cases, \cite{garcia2021assessment} applied an RF model to a cohort of 1,525 patients. In \cite{zamanian2024estimation}, a comparative analysis of multiple ML methods, including LR, Linear Discriminant Analysis, RF, AdaBoost, KNN, SVM, Multilayer Perceptron (MLP), and DT, was conducted on 181 patients. Ma et al. \cite{ma2018application} applied a BN on a sample of 10,508 subjects to diagnose NASH. Authors of \cite{docherty2021development} and \cite{schattenberg2023nashmap} classified subjects with or without NASH using XGBoost. Although numerous articles developed ML models based on biopsy \cite{forlano2020high}, ultrasound images \cite{destrempes2022quantitative}, or Dual-energy X-ray absorptiometry features \cite{boncan2024machine} for staging liver steatosis, there was only one paper that utilized clinical data for this purpose. Razmpour et al. \cite{razmpour2023application}] utilized various ML methods, including KNN, SVM, Radial Basis Function (RBF) SVM, Gaussian Process (GP), RF, NN, AdaBoost, and NB to classify NAFLD based on body composition and anthropometric measurements.

Their results indicated that the RF generated the most accurate model for staging the steatosis.
These studies showcase various ML techniques and features for classifying NAFLD and its related conditions. Common classifiers such as XGBoost and RF are frequently employed, utilizing features ranging from primary demographic data to detailed clinical and biochemical markers. The sample sizes in these studies vary significantly, affecting the models' generalizability and robustness. These studies underscore both the potential and challenges of applying ML in the clinical diagnostics of liver diseases.
In summary, the primary benefits of the suggested approach are:

\begin{enumerate}
  \item The novel stacking model uses information fusion techniques at data and feature levels, improving predictive accuracy and robustness.
  \item Classifiers in each stacking layer are chosen based on test performance, achieving 86\% accuracy and a 96\% AUC, making it more reliable than single classifiers due to its use of multiple models and information fusion.
  \item The model’s performance is thoroughly evaluated using various metrics and 10-fold CV, ensuring a comprehensive assessment.
  \item By integrating multiple classifiers, the model addresses individual model weaknesses, resulting in a more robust system that performs well even with incomplete data.
  \item The architecture allows flexibility in incorporating different classifiers and features, making it adaptable to various datasets and clinical scenarios.
  \item The ensemble method effectively handles missing data, making it highly useful in clinical settings where incomplete datasets are common.
\end{enumerate}

However, several limitations should be addressed in future studies:

\begin{enumerate}
  \item The multi-layer stacking model increases computational complexity, and future work could optimize it to reduce resource demands without losing performance.
  \item Further validation of external datasets is needed to confirm their generalizability across different populations.
  \item The model's complexity may hinder interpretability, so future research should focus on enhancing its transparency for clinical use.
  \item Expanding the model to assess fibrosis stages and diagnose early-stage hepatocellular carcinoma in NASH patients would be beneficial.
\end{enumerate}

\section {Conclusion}
This study underscores the importance of developing non-invasive and reliable diagnostic models for NAFLD, notably NASH, to address the limitations of current diagnostic methods. By integrating clinical data and laboratory test results into machine learning models, the proposed stack-based ensemble classifier improves accuracy and robustness in diagnosing various stages of NAFLD. The cascade model effectively mitigates the issues associated with missing data through ensemble learning and information fusion techniques, making it a valuable tool for clinical and epidemiological applications. This development shows potential for improving patient outcomes by enabling early diagnosis and management of NAFLD, minimizing the need for invasive procedures, and enhancing overall disease monitoring and treatment approaches.

\section*{References}
\bibliographystyle{IEEEtran}
\bibliography{references}

\end{document}